\documentclass[letterpaper, 10 pt, journal, twoside]{IEEEtran}
\ifCLASSINFOpdf
\else
\fi
\usepackage{graphicx}
\usepackage{xcolor}
\usepackage{amsmath,amssymb} 

\newcommand{\minunder}{\mathop{\mathrm{min}}\limits}
\newcommand{\maxunder}{\mathop{\mathrm{max}}\limits}

\newcommand{\argmin}{\mathop{\mathrm{arg min}}\limits}

\usepackage{amssymb}
\usepackage{multirow}
\usepackage{booktabs}
\usepackage{mathtools,cuted}
\usepackage{subcaption}
\usepackage[compress]{cite}

\begin{document}
%
\title{Multi-View Object Pose Refinement with Differentiable Renderer}
%
%
%

\author{Ivan Shugurov$^{*,1,2}$, Ivan Pavlov$^{*,1,2}$, Sergey Zakharov$^{1,2,3}$ and Slobodan Ilic$^{1,2}$
\thanks{Manuscript received: October, 15, 2020; Revised: January, 14, 2021; Accepted: February, 8, 2021.}%
\thanks{This paper was recommended for publication by Editor Markus Vincze upon evaluation of the Associate Editor and Reviewers' comments.)} %
\thanks{$^{1}$Department of Informatics, Technical University of Munich, Germany
        {\tt\small ivan.shugurov@tum.de}}%
\thanks{$^{2}$Siemens AG, Munich, Germany}%
\thanks{$^{3}$Currently at Toyota Research Institute, Los Altos, USA}%
\thanks{$^{*}$These authors contributed equally to the work.}%
\thanks{Digital Object Identifier (DOI): 10.1109/LRA.2021.3062350}
}
%
%

\markboth{IEEE Robotics and Automation Letters. Preprint Version. ACCEPTED February, 2021}
{Shugurov \MakeLowercase{\textit{et al.}}: Multi-View Pose Refinement with Differentiable Renderer} 

%



\maketitle

\begin{abstract}
This paper introduces a novel multi-view 6 DoF object pose refinement approach focusing on improving methods trained on synthetic data. It is based on the DPOD detector, which produces dense 2D-3D correspondences between the model vertices and the image pixels in each frame. We have opted for the use of multiple frames with known relative camera transformations, as it allows introduction of geometrical constraints via  an interpretable ICP-like loss function. The loss function is implemented with a differentiable renderer and is optimized iteratively. We also demonstrate that a full detection and refinement pipeline, which is trained solely on synthetic data, can be used for auto-labeling real data. We perform quantitative evaluation on LineMOD, Occlusion, Homebrewed and YCB-V datasets and report excellent performance in comparison to the state-of-the-art methods trained on the synthetic and real data. We demonstrate empirically that our approach requires only a few frames and is robust to close camera locations and noise in extrinsic camera calibration, making its practical usage easier and more ubiquitous.
\end{abstract}


%
\IEEEpeerreviewmaketitle

\section{Introduction}

\IEEEPARstart{O}{bject} detection and 6D pose estimation in RGB images are among the most fundamental problems in computer vision, with applications encompassing autonomous driving, augmented reality and robotics. A large body of work has already been presented, but recent advances in deep learning have opened up new horizons for RGB-based algorithms, which have now started to dominate the field. However, precise 6 DoF pose estimation, required, in autonomous robotic grasping systems, for example, still remains a challenging problem. This is mainly due to the perspective ambiguity, lightning changes, clutter and occlusions. Current industrial implementations~\cite{drost2010model} are not based on deep learning and rely on depth data for increased pose accuracy. Moreover, these approaches use 3D models directly, while deep learning methods still struggle with the domain gap when trained on synthetic data rendered from 3D CAD models. Recently, CosyPose~\cite{labbe2020}, a deep learning method that operates on RGB images, assumed the lead over traditional methods in the BOP challenge\cite{bopchallenge}, especially when multiple images are used for pose refinement.

The task of pose refinement has also been addressed by deep learning approaches. Methods like ~\cite{zakharov2019dpod,manhardt2018deep,li2018deepim} trained CNN networks on pairs of images, the aim being
to learn to predict the pose offset between the predicted object pose and the observed detected object. There, 3D models were rendered in the predicted pose and real images were given as patches with the detected object. The main disadvantage of these refiners is that the are dependent on the correct pose error priors used during training coupled with the necessity to re-train them for each new object in order to obtain high-quality results. 

We address the aforementioned problems in this paper by introducing a multi-view refinement procedure. We first train the DPOD detector~\cite{zakharov2019dpod} on synthetic data to overcome the dataset bias and the lack of real training data. We then exploit geometric multi-view constraints during refinement to cope with perspective ambiguities and occlusions in monocular RGB images that cause their sub-par performance.  Compared to existing deep learning refiners ~\cite{zakharov2019dpod,manhardt2018deep,li2018deepim}, the proposed approach has the following advantages: 1) it does not require training of the refiner itself;  2) it is not object-specific;  3) it can be applied to arbitrary many images without the need for any modification;  4) it has explicit geometric constraints and, thus, an explicit objective function which is optimized during the refinement.

\begin{figure*}[!t]
	\centering
	\includegraphics[width=0.7\textwidth]{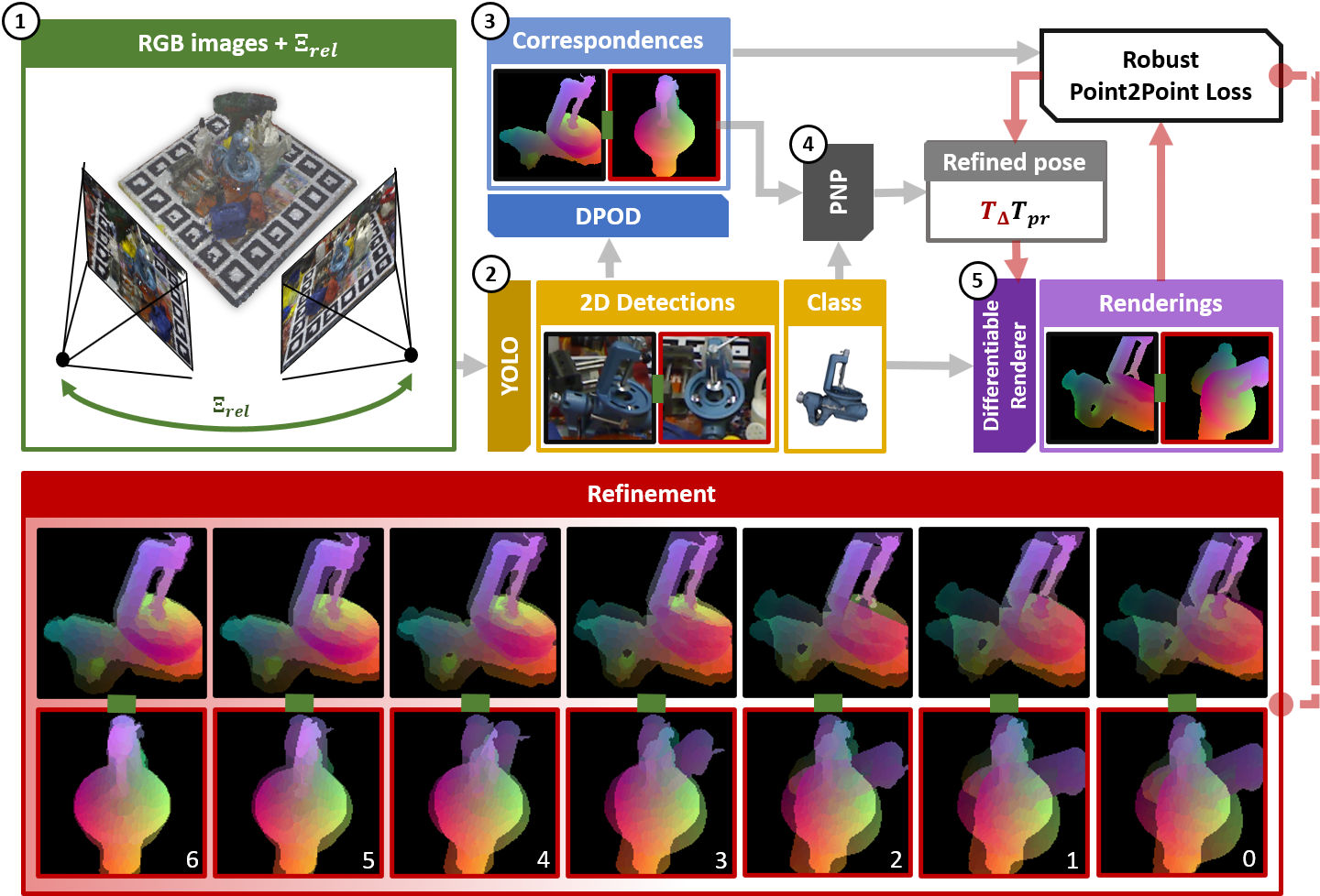}
	\caption{\textbf{Multi-view inference and pose optimization}. 1) Inputs to the algorithm  are an unordered set of images and corresponding relative camera transformations. 2) YOLO is applied to each image separately to detect the object of interest in each of them. 3) Dense correspondences are predicted with the DPOD network. 4) Rough object pose in the reference frame is estimated using PnP and RANSAC using the predicted 2D-3D dense correspondences. 5) The final pose is iteratively refined using the multi-view optimization based on differentiable rendering.}
	\label{fig:training}  
	\vspace{-1.5em}
\end{figure*}

The proposed pose refinement procedure is based on differentiable renderering. It uses multiple views  to add relative camera poses as constraints to the pose optimization procedure. It is done by comparing predicted and rendered dense correspondences in each frame and then transmitting the error back though the differentiable renderer to update the pose. The proposed loss function allows varying numbers of frames to be used without any changes to the optimization procedure. The loss formulation also does not impose any restrictions on where in the 3D world the cameras are placed and whether or not views from different cameras overlap as long as the object is visible in the images. We assume the availability of relative camera poses. In practice, they can be easily obtained by a number of various methods, such as placing the object on the markerboard and either using an actual multi-camera system or using a single camera but moving the markerboard.  In the scenario of robotic grasping, a camera can be mounted on the robotic arm to enable observation of the object from several viewpoints.


We evaluate this approach on LineMOD~\cite{hinterstoisser2012model}, Occlusion~\cite{brachmann2014learning}, Homebrewed~\cite{Kaskman_2019_ICCV_Workshops} and YCB-V~\cite{xiang2018posecnn} datasets and report performance that is superior to any related method trained on synthetic data and similar to or better than methods which use real training data and post-refinement. Our experiments show that our approach can robustly perform multi-view pose refinement even when relative poses are imprecise.  Our results also demonstrate that the proposed refinement remains effective even in degenerate cases, when cameras are in close proximity to each other. 

Further, we show that our framework can be used for the task of auto-labeling real data as in~\cite{zakharov2020autolabeling}, thus removing the need for manual pose labeling. The networks are first trained on synthetic data and then used to label the real images. 
This procedure enables a considerable reduction in the time and effort needed to annotate the data.  Our multi-view refinement pipeline enables us to automatically produce high quality pose annotations for these real images and re-train the detector on them.


\section{Related Work}

This section first discusses the state of the art of monocular pose estimation, before going on to review popular pose refinement techniques and discuss methods that use multiple images for object detection and pose estimation. 

\begin{figure*}[!t]
\begin{subfigure}{.32\textwidth}
  \centering
  \includegraphics[width=.48\linewidth]{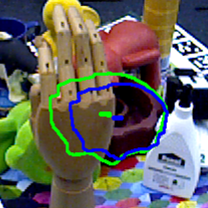}
  \hfill
  \includegraphics[width=.48\linewidth]{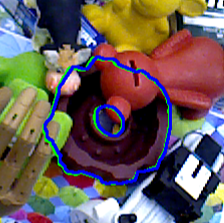}
  \hfill
  
  \hphantom{}
   
  \includegraphics[width=.48\linewidth]{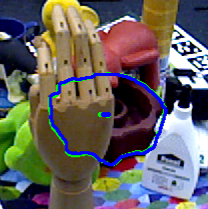}
  \hfill
  \includegraphics[width=.48\linewidth]{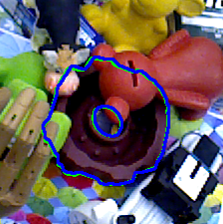}
  \hfill
  
  \caption{Refinement with two views.}
  
\end{subfigure}%
\hfill
\begin{subfigure}{.64\textwidth}
  \centering
  \hfill
  \includegraphics[width=.24\linewidth]{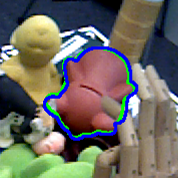}
  \hfill
  \includegraphics[width=.24\linewidth]{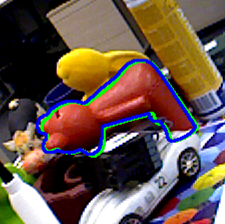}
  \hfill
  \includegraphics[width=.24\linewidth]{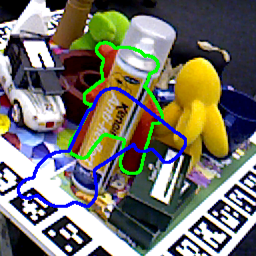}
  \hfill
  \includegraphics[width=.24\linewidth]{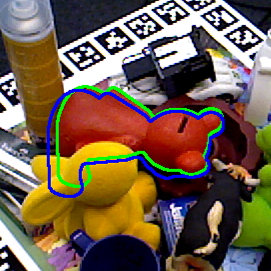}
  
  \hphantom{}
   
   \hfill
  \includegraphics[width=.24\linewidth]{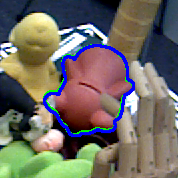}
  \hfill
  \includegraphics[width=.24\linewidth]{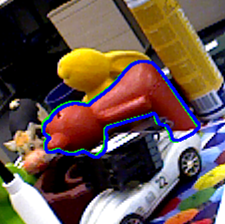}
  \hfill
  \includegraphics[width=.24\linewidth]{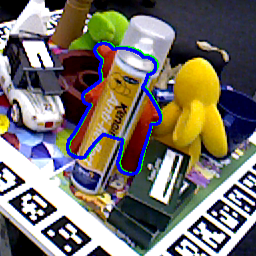}
  \hfill
  \includegraphics[width=.24\linewidth]{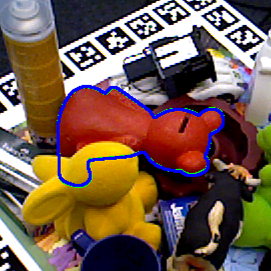}
  
  \caption{Refinement with four views.}
\end{subfigure}
\caption{Example refinement results on the Homewbrewed dataset~\cite{Kaskman_2019_ICCV_Workshops}. The top row shows initial per-frame poses produced by PnP before refinement, while the bottom row shows them after refinement. The outline of the object is visualized in green for the ground truth pose, and in blue for the estimated pose. This illustrates that the proposed refiner is capable of selecting a reference frame with a good initial pose and refining it even in the presence of occlusions and imprecise correspondences or when some of the initial pose hypotheses are completely incorrect, as in (b).  \label{fig:refinement_vis}}
  \vspace{-2.0em}
\end{figure*}

\textbf{6 DoF pose estimation} SSD-6D~\cite{kehl2017ssd} was one of the first deep learning methods for 6 DoF pose estimation from RGB images. SSD-6D extends the off-the-shelf SSD object detector and was fully trained on synthetic data. Since the method is based on learning discrete poses, it 
requires refinement to obtain reasonable results.
Subsequent approaches predicted a fixed number of 2D keypoints, whose locations on the 3D model are known. Using those 2D-3D correspondences, the full pose is estimated using a PnP algorithm. BB8\cite{rad2017bb8} proposed predicting 2D projections of the 3D bounding box corners. In contrast, PVNet~\cite{peng2019pvnet} used per-pixel Hough voting to allow all object's pixels to vote for the few keypoints lying on the object. HybridPose~\cite{song2020hybridpose} extended correspondence prediction by predicting edge vectors and symmetry correspondences, which were then used jointly in an augmented PnP algorithm. The concept of 2D-3D correspondence estimation was further extended in the methods that rely on dense correspondences: DPOD~\cite{zakharov2019dpod}, Pix2Pose~\cite{park2019pix2pose}, CDPN~\cite{li2019cdpn} and EPOS~\cite{Hodan_2020_CVPR}. They predicted a 3D correspondence for each foreground object pixel, thus increasing the number of correspondences. AAE~\cite{sundermeyer2018implicit} used an off-the-shelf 2D object detector trained on real data and an autoencoder, trained on sythetic data, followed by feature matching to predict discrete object poses. PoseCNN~\cite{xiang2018posecnn} relied on Hough voting to locate the 2D projection of object's center and distance from the object to estimate the translation. Rotation was directly regressed as a quaternion.

Of the aforementioned methods, only SSD6D, DPOD, AAE and EPOS reported results on synthetic train data in the original papers, though for some of them the results on synthetic data were later submitted to the BOP challenge~\cite{bopchallenge}. 

\textbf{6D pose refinement.} Object pose refinement has been studied thoroughly in the past. Iterative Closest Point~\cite{icp} (ICP) is one of the most classical and used approaches. 
ICP refines a given pose iteratively by establishing one-to-one correspondences between the point cloud and object's vertices based on their spatial proximity and then minimizing distances between them. There are various variants of ICP and ICP-like algorithms, but their common fundamental limitation is that they require reliable registered depth information, which is not always readily available. 
It restricts their applicability in many scenarios, for example, on images from mobile phones or from edge devices.
Another line of research attempts to replace depth-based ICP with deep learning on point clouds. The most relevant example is DenseFusion~\cite{wang2019densefusion}, which first applies a convolutional neural network to segment out the objects of interests. It then fuses RGB and depth features to predict the pose. A trainable iterative deep refiner is applied subsequently to directly regress the pose offset. 

In RGB images, the object pose has been traditionally refined using edge alignment. The idea is to align the edges of the object rendered in the pose hypothesis with the edges observed in the image. This approach is very sensitive to the image quality, clutter, occlusions and to the way in which the edge correspondences are computed. 
Later, deep learning pose refinement approaches~\cite{zakharov2019dpod,manhardt2018deep,li2018deepim} operating on RGB images were introduced. The idea here is to use an external pose estimation algorithm to obtain the initial pose approximation. Then, the object is rendered in the predicted pose. A rendered image and a given input RGB image are then put into another convolutional neural network, which directly regresses a pose offset. In BB8~\cite{rad2017bb8}, on the other hand, the refiner was trained to update predicted locations of keypoints.  

Disadvantages of the aforementioned deep learning-based refiners are their dependence on the correct pose error priors used during training and the need to retrain them for each new object in order to obtain high-quality results. 

\textbf{Object pose estimation and refinement from multiple views.} This topic has been studied in the past in several works, most notably in~\cite{sock2017multi,erkent2016integration,li2018unified}. The papers display two main trends. First, they all rely on independent pose hypothesis prediction for each monocular image. Secondly, relative camera transformations between frames are assumed to be known beforehand. Known camera transformations are used to fuse pose predictions from several views in the global coordinate system. Then, either the pose hypothesis that aligns the best with the other hypotheses is chosen or poses are refined to align better in the 3D space.
Orthogonal to those methods, are the approaches proposed in~\cite{kaskman2020,labbe2020}. In~\cite{kaskman2020}, a set of uncalibrated RGB images is used to create a scale-ambiguous 3D reconstruction. 
Then, objects are detected in the 3D reconstruction that correspond to joint detection in all the frames. Despite its good performance, the method had the downside that it comprised a number of complicated and time-consuming steps and the need to use a large number (72) of frames to obtain reliable reconstructions. CosyPose~\cite{labbe2020} also relies on independent pose hypotheses from each frame. However, in contrast to~\cite{sock2017multi,erkent2016integration,li2018unified}, they are then used in a RANSAC scheme to match pose hypotheses from several frames and produce a unified object-level scene reconstruction and approximate relative camera poses. They are jointly optimized by minimizing the multi-view reprojection error. Unfortunately, each of these methods is evaluated on different datasets, and different pose metrics are reported, which makes it impossible to compare them either with each other or with the recent pose estimation papers.  We compare our refiner to CosyPose on Homebrewed~\cite{Kaskman_2019_ICCV_Workshops} and YCB~\cite{xiang2018posecnn} datasets.

\section{Proposed Method}

The complete inference pipeline is shown in Figure~\ref{fig:training}. The proposed method is divided into the following steps. Step 1: takes a set of images together with their known intrinsic parameters and relative transformations between cameras; Step 2: The objects of interest in each image are detected separately; Step 3: The per-object per-pixel 2D-3D correspondences are predicted independently for each detected object; Step 4: The object 6 DoF pose is estimated with EPnP~\cite{lepetit2009epnp} and RANSAC using the predicted 2D-3D correspondences; Step 5: The initial rough pose is iteratively refined to find a pose which better aligns with predicted dense correspondences in all the frames. This is done by defining a loss function over the predicted correspondences and the ideal 2D-3D correspondences which correspond to the object in the given pose. These ideal correspondences are produced with a differentiable renderer so that the entire multi-view alignment procedure is differentiable. We will now describe each step in more detail. However, the main contribution of the paper comprises the multi-view refinement in Step 5. In all experiments, we used the Soft Rasterizer renderer~\cite{liu2019soft}.

\subsection{Object Detection and Pose Estimation}
As the focus of the paper is on refinement and not on the whole pipeline, we have opted for one of the already available dense correspondence-based detectors. We slightly extended the DPOD~\cite{zakharov2019dpod} and trained it on synthetic training data (later referred to as PBR data) provided by the BOP challenge~\cite{bopchallenge}. We separated the original DPOD architecture into two parts: 1) the YOLOv3~\cite{redmon2018yolov3} detector trained to output tight object bounding boxes with corresponding semantic labels, and 2) the DPOD-like architecture, which predicts object masks and dense correspondences from these detections.
The proposed two-stage approach simplifies and accelerates the training procedure of each component, slightly improves the quality of correspondences and leads to better performance on more challenging Homebrewed~\cite{Kaskman_2019_ICCV_Workshops} and YCB-V\cite{xiang2018posecnn} datasets.
In contrast to DPOD~\cite{zakharov2019dpod}, which relied on two-dimensional UV maps, we use the three-dimensional Normalized Object Coordinates Space (NOCS)~\cite{wang2019normalized}. 
This parameterization permits trivial conversion between the object coordinate system and the NOCS coordinate system. 
Additionally, we switched from the ResNet18 backbone to the ResNet34 backbone. The first block of layers is frozen to avoid overfitting when training on synthetic data. 
Data augmentation and transfer learning allowed the reliable training of the networks.  YOLO was trained for 100 epochs, the augmented DPOD for 240. The last checkpoint was used in all experiments.

Let us formally define a model as a set of its vertices: $\mathcal{M} \coloneqq \{v \in \mathbb{R}^3\}$. Operators that compute minimum and maximum coordinates along the vertex dimension $i$ of all $v \in \mathcal{M}$ are defined as:
\begin{equation}
\begin{aligned}[c]
min_i(\mathcal{M}) \coloneqq \minunder_{v \in \mathcal{M}}v_i,
\end{aligned}
\qquad\qquad
\begin{aligned}[c]
max_i(\mathcal{M}) \coloneqq \maxunder_{v \in \mathcal{M}}v_i
\end{aligned}
\label{eq:min_max_operator}
\end{equation}

Then, for any point $p$, the NOCS projection operator is defined w.r.t. the model as 
\begin{equation}
    \pi_{\mathcal{M}}(p) \coloneqq \left\{ \frac{p_d - min_d(\mathcal{M})}{max_d(\mathcal{M})-min_d(\mathcal{M})}\right\}_{d \in \left\{x, y, z\right\}}
    \label{ea:nocs}
\end{equation}

 and its inverse as $\pi_{\mathcal{M}}^{-1}$. The 6 DoF pose is defined as the standard rigid body transformation, where $\mathbf{T}\in  \mathcal{SE}(3)$.

\subsection{Pose refinement with differentiable renderer}

Given a ground truth pose $\mathbf{T}_{gt}$ and a predicted pose $\mathbf{T}_{pr}$, the aim of pose refinement is to find a pose update $\mathbf{T}_\Delta$ that satisfies  
$\mathbf{T}_\Delta \cdot \mathbf{T}_{pr} = \mathbf{T}_{gt}$.
It is, however, impossible to estimate a perfect $\mathbf{T}_\Delta$, because the ground truth pose  $\mathbf{T}_{gt}$ is not available. For this reason, proxy loss functions have to be used, which results in a sub-optimal pose update $\mathbf{T}_\Delta$. We propose refining $\mathbf{T}_{pr}$ by optimizing the discrepancy between the predicted noisy NOCS maps from several frames and the perfect NOCS map rendered in the estimated pose with a differentiable renderer. 

\begin{table*}[!t]
  \centering
  \caption{Percentages of correctly estimated poses w.r.t. the ADD on the LineMOD~\cite{hinterstoisser2012model} dataset. DR1, DR2 and DR4 stand for the proposed refinement with 1, 2 and 4 views respectively. Single-asterisked (*) methods use real training data. Double-asterisked (**) methods report only refinement time instead of the time of the whole pipeline.
  \label{tab:lm}}
  \resizebox{1\linewidth}{!}{%
       \begin{tabular}{c|cccccc|cc|cc|cc|cccc}
\cmidrule{2-17}          & \multicolumn{6}{c|}{\multirow{2}[4]{*}{\textbf{RGB Single-View}}} & \multicolumn{6}{c|}{\textbf{RGB Multi-View}}  & \multicolumn{4}{c}{\multirow{2}[4]{*}{\textbf{RGBD}}} \\
\cmidrule{8-13}          & \multicolumn{6}{c|}{}                         & \multicolumn{2}{c|}{\textbf{Closest views}} & \multicolumn{2}{c|}{\textbf{Random views}} & \multicolumn{2}{c|}{\textbf{Farthest views}} & \multicolumn{4}{c}{} \\
\cmidrule{2-17}    \textbf{Method} & \textbf{SSD6D\cite{kehl2017ssd}} & \textbf{OURS} & \textbf{BB8*\cite{rad2017bb8}} & \textbf{DPOD\cite{zakharov2019dpod}} & \textbf{OURS} & \textbf{PoseCNN*\cite{xiang2018posecnn}} & \multicolumn{2}{c|}{\textbf{OURS}} & \multicolumn{2}{c|}{\textbf{OURS}} & \multicolumn{2}{c|}{\textbf{OURS}} & \textbf{AAE\cite{sundermeyer2018implicit}} & \textbf{SSD6D\cite{kehl2017ssd}} & \textbf{DenFus*\cite{wang2019densefusion}} & \textbf{DenFus.*\cite{wang2019densefusion}} \\
    \textbf{Refinement} & \textbf{ DL\cite{manhardt2018deep}} & \textbf{-} & \textbf{DL\cite{rad2017bb8}} & \textbf{ DL\cite{zakharov2019dpod}} & \textbf{ DR1} & \textbf{DeepIM\cite{li2018deepim}} & \textbf{DR2} & \textbf{DR4} & \textbf{DR2} & \textbf{DR4} & \textbf{DR2} & \textbf{DR4} & \textbf{ ICP} & \textbf{ ICP} & \textbf{-} & \textbf{DL\cite{wang2019densefusion}} \\
    \midrule
    Ape   & -     & 28.04 & 40.40 & 55.23 & 38.351 & 76.95 & 55.28 & 85.38 & 97.89 & \textbf{100}   & 97.7  & \textbf{100}   & 24.35 & -     & 80    & 92 \\
    Bvs.  & -     & 71.65 & 91.80 & 72.69 & 85.354 & 97.48 & 94.95 & 99.61 & 99.81 & \textbf{100}   & \textbf{100}   & 99.61 & 89.13 & -     & 84    & 93 \\
    Cam   & -     & 58.62 & 55.70 & 34.76 & 76.399 & 93.53 & 87.57 & 98.18 & 99.7  & 100   & 99.7  & \textbf{100}   & 82.1  & -     & 77    & 94 \\
    Can   & -     & 62.8  & 64.10 & 83.59 & 87.316 & 96.46 & 89.57 & 96.06 & 95.47 & 96.06 & 95.87 & \textbf{96.85} & 70.82 & -     & 87    & 93 \\
    Cat   & -     & 40.12 & 62.60 & 65.1  & 55.988 & 82.14 & 78.4  & 97.2  & 99.4  & \textbf{100}   & 99.4  & \textbf{100}   & 72.18 & -     & 89    & 97 \\
    Driller & -     & 61.46 & 74.40 & 73.32 & 79.801 & 94.95 & 86.5  & 96.02 & 97.13 & 97.79 & \textbf{98.23} & 97.79 & 44.87 & -     & 78    & 87 \\
    Duck  & -     & 26.17 & 44.30 & 50.04 & 38.18 & 77.65 & 53.57 & 81.2  & 94.56 & \textbf{98.87} & 97.56 & \textbf{98.87} & 54.63 & -     & 76    & 92 \\
    Eggbox & -     & 87.65 & 57.80 & 89.05 & 93.003 & 97.09 & 95.53 & 98.83 & 97.27 & 99.22 & 98.83 & 99.61 & 96.62 & -     & \textbf{100}   & \textbf{100} \\
    Glue  & -     & 67.48 & 41.20 & 84.37 & 67.988 & 99.42 & 89.15 & 97.67 & 95.94 & 98.06 & 93.22 & 98.06 & 94.18 & -     & 99    & \textbf{100} \\
    Holep. & -     & 19.09 & 67.20 & 35.35 & 24.568 & 52.81 & 35.38 & 52.69 & 79.27 & 87.69 & 84.81 & 87.31 & 51.25 & -     & 79    & \textbf{92} \\
    Iron  & -     & 83.85 & 84.70 & 98.78 & 94.28 & 98.26 & 98.36 & \textbf{100}   & \textbf{100}   & \textbf{100}   & \textbf{100}   & \textbf{100}   & 77.86 & -     & 92    & 97 \\
    Lamp  & -     & 43.52 & 76.50 & 74.27 & 65.664 & 97.5  & 80.38 & 91.05 & 94.37 & 97.28 & 95.92 & \textbf{96.5}  & 86.31 & -     & 92    & 95 \\
    Phone & -     & 45.77 & 54.00 & 46.98 & 74.472 & 87.72 & 87.32 & 97.18 & 99.6  & \textbf{100}   & 99.4  & 99.6  & 86.24 & -     & 88    & 93 \\
    \midrule
    Mean  & 34.1  & 53.55 & 62.70 & 66.42 & 67.79 & 88.61 & 79.38 & 91.62 & 96.19 & 97.19 & 96.97 & \textbf{98.02} & 71.58 & 90.9  & 86    & 94 \\ \midrule
    Time (ms) & -     & 56    & 330   & 36    & 398   &  + 83** & 233   & 159   & 232   & 162   & 227   & 163   & 224   & 100   & -     & - \\
    \end{tabular}%
    }
    \vspace{-1.5em}
\end{table*}%

Assuming that N frames are used for multi-view refinement, we define the corresponding set of predicted segmentations $\tilde{\mathcal{S}} \coloneqq  \{\tilde{\mathbf{S}}_1, \dots, \tilde{\mathbf{S}}_N \}$, each of which is a binary segmentation mask $\tilde{\mathbf{S}} \in \{0, 1\}^{W \times H}$.  A set of predicted NOCS correspondences is defined as $\tilde{\mathcal{C}} \coloneqq \{\tilde{\mathbf{C}}_1, \dots, \tilde{\mathbf{C}}_N\}$, where each $\tilde{\mathbf{C}} \in [0, 1]^{W\times H \times3}$. These are noisy estimates obtained with the modified DPOD. They are computed once and remain unchanged during refinement. The differentiable renderer is used for a differentiable definition of the following functions: binary foreground/background rendering $S: \mathcal{M} \times \mathcal{SE}(3) \rightarrow \{0, 1\}^{W \times H}$ and NOCS rendering $C: \mathcal{M} \times \mathcal{SE}(3) \rightarrow [0, 1]^{W \times H \times 3}$. We will omit their dependence on the CAD model $\mathcal{M}$ to keep the notation concise. For each set of images used for multi-view refinement, one of the images is taken as the reference frame, and then for each $f$-th image,  its relative pose $\mathbf{\Xi}_{ref\rightarrow f} \in \mathcal{SE}(3)$ is recomputed w.r.t. to the reference frame. Initial pose hypotheses $\mathcal{T} \coloneqq \left\{ \mathbf{T}_1, ..., \mathbf{T}_N \right\}$ are estimated separately for each frame with PnP+RANSAC. The pose of the object in the coordinate system of the reference frame is later denoted by $\mathbf{T}_{pr}$.

We use pre-computed NOCS maps in each frame and NOCS maps of the model in the estimated pose in the reference frame transformed to the coordinate system of the $f$-th frame by transformation $\mathbf{\Xi}_{ref\rightarrow f} \cdot \mathbf{T}_{pr}$ to define a per-pixel loss function over predicted and rendered correspondences. The per-pixel NOCS discrepancy relates directly to the 3D structure of the object and the error in 3D. The per-pixel loss for pixel $p$ in the $f$-th frame is defined as follows:
\begin{equation}
    \mathcal{L}_{f, p} \coloneqq \rho\left(\pi^{-1}\left(\tilde{\mathbf{C}}_{f,p}\right), \pi^{-1}\left(C\left(\mathbf{\Xi}_{ref\rightarrow f} \cdot \mathbf{T}_{\Delta} \cdot \mathbf{T}_{pr}\right)_p\right)\right)
\end{equation}

In the above equation, $\rho$ stands for the arbitrary distance function in 3D. Essentially, for each frame, the object pose $\mathbf{T}_{\Delta} \cdot \mathbf{T}_{pr}$ is first transformed to the frame's coordinate system using $\mathbf{\Xi}_{ref\rightarrow f}$ and then rendered. For each pixel, the predicted and rendered NOCS correspondences are projected into the 3D coordinate system of the model. Then, the discrepancy between them is penalized. The overall loss function is fully differentiable and dependent only on $\mathbf{T}_{\Delta}$, since the rendering of NOCS maps is performed using the differentiable renderer. 

On the level of the full set of images used for refinement, the objective of the refinement is defined as:
\begin{equation}
\resizebox{1\linewidth}{!}{
    $\mathbf{T}_\Delta^{*} = \argmin_{\mathbf{T}_\Delta \in \mathcal{SE}(3)} \sum_{f=1}^{N} \sum_{p \in \mathcal{I}} \left[\tilde{\mathbf{S}}_{f, p} \cdot S\left(\mathbf{\Xi}_{ref\rightarrow f} \cdot \mathbf{T}_{\Delta} \cdot \mathbf{T}_{pr}\right)_p\right] \cdot \mathcal{L}_{f, p}$
    }
\end{equation}
Here, $\left[\tilde{\mathbf{S}}_{f, p} \cdot S\left(\mathbf{\Xi}_{ref\rightarrow f} \cdot \mathbf{T}_{\Delta} \cdot \mathbf{T}_{pr}\right)_p\right]$ serves as an indicator function regarding whether or not the pixel is foreground in both the rendered and the predicted NOCS maps.

The minimization problem, however, cannot be solved optimally due to its non-convexity. Therefore, the loss function is minimized iteratively by gradient descent over the pose update $\mathbf{T}_\Delta$. This can be done using any gradient-based method. We normally observe convergence within 50 optimization steps. A trivial degenerate solution to the optimization problem exists, which sets the loss to zero, namely non-overlapping rendered and predicted  $\tilde{\mathcal{S}}$ segmentation maps. But in reality, this is not a problem, because PnP+RANSAC provides reliable initial pose estimates. Moreover, degenerate pose hypotheses are explicitly handled in the choice of the reference frame.

There are numerous ways os implementing the distance function $\rho \colon \mathbb{R}^3 \times \mathbb{R}^3 \to \mathbb{R}^{+}$ and parameterizing $\mathcal{SO}(3)$ rotations. We use the continuous rotation parameterization from~\cite{zhou2019continuity}, which enables faster and more stable convergence during the optimization procedure than quaternions and Euler angles. As the predicted correspondences and matched correspondences might contain a potentially large number of outliers, a robust $\rho$ function must be used to mitigate this. We experimented with several options and ended up with a particular case of the general robust function introduced in~\cite{barron2019general}. This function is defined as follows:
\begin{equation}
    \rho(e, c) \coloneqq 1 - \exp{\left( -\frac{1}{2} \left(\frac{e}{c}\right)^2 \right)}
\end{equation}
$c$ is a hyper-parameter that stands for the scale of the loss function. In our experiments, we adjusted it dynamically according to the median absolute residuals:  $c \coloneqq 2 \cdot MEDIAN\left(\left|\mathbf{e}\right|\right)$.

The choice of the reference frame can affect the effectiveness of pose refinement. The goal is to automatically choose a pose which has a high overlap with predicted segmentations when transformed and rendered in other views. Additionally, it should have the smallest possible loss $\mathcal{L}_f$. For each image batch, the reference frame is chosen as follows:
\begin{equation}
   \argmin_{ref \in \left[1, .., N\right]} \frac{1}{K} \sum_{f=1}^N 
    \frac{
    \sum_{p \in \mathcal{I}} \left[\tilde{\mathbf{S}}_{f, p} \cdot S\left(\mathbf{\Xi}_{ref\rightarrow f} \cdot \mathbf{T}_{ref}\right)_p\right] \cdot \mathcal{L}_{f, p}
    }
    {
    IOU\left(\tilde{\mathbf{S}}_{f}, S\left(\mathbf{\Xi}_{ref\rightarrow f} \cdot \mathbf{T}_{ref}\right)\right)
    }
\end{equation}
where $K$ stands for the number of frames with non-zero loss. Additionally, $\argmin$ ignores zero values, as they correspond to degenerate poses that are not re-projected correctly onto other frames.

\begin{table*}[t]
  \centering
  \caption{Percentages of correctly estimated poses w.r.t. the ADD on the Occlusion~\cite{brachmann2014learning} dataset. DR1, DR2 and DR4 stand for the proposed refinement with 1, 2 and 4 views respectively.
  \label{tab:occlusion}}
  \resizebox{1\linewidth}{!}{%
    \begin{tabular}{c|ccc|cc|cc|cc|cccc}
    \textbf{Train data} & \multicolumn{9}{c|}{\textbf{Synthetic}}                               & \multicolumn{4}{c}{\textbf{Real - GT labels}} \\
    \midrule
    \multirow{2}[4]{*}{\textbf{Method}} & \multirow{2}[4]{*}{\textbf{OURS}} & \multirow{2}[4]{*}{\textbf{SSD6D}\cite{kehl2017ssd}} & \multirow{2}[4]{*}{\textbf{OURS}} & \multicolumn{6}{c|}{\textbf{OURS}}          & \multirow{2}[4]{*}{\textbf{Pix2Pose}\cite{park2019pix2pose}} & \multirow{2}[4]{*}{\textbf{PVNet}\cite{peng2019pvnet}} & \multirow{2}[4]{*}{\textbf{DPOD}\cite{zakharov2019dpod}} & \multirow{2}[4]{*}{\textbf{HybPose}\cite{song2020hybridpose}} \\
\cmidrule{5-10}          &       &       &       & \multicolumn{2}{c|}{\textbf{closest views}} & \multicolumn{2}{c|}{\textbf{random views}} & \multicolumn{2}{c|}{\textbf{farthest views}} &       &       &       &  \\
    \midrule
    \textbf{Refinement} & \textbf{-} & \textbf{DL} \cite{manhardt2018deep} & \textbf{DR1} & \textbf{DR2} & \textbf{DR4} & \textbf{DR2} & \textbf{DR4} & \textbf{DR2} & \textbf{DR4} & \textbf{-} & \textbf{-} & \textbf{-} & \textbf{-} \\
    \midrule
    Ape   & 26.1  & -     & 15.9  & 21.59 & 31.45 & 57.85 & \textbf{79.81} & 61.74 & \textbf{79.81} & 22.0  & 15.8  & 18.8  & 53.3 \\
    Can   & 37.8  & -     & 45.4  & 63.61 & 81.67 & 71.57 & \textbf{88.05} & 72.37 & 87.65 & 44.7  & 63.3  & 60.8  & 86.5 \\
    Cat   & 6.3   & -     & 8.3   & 20.45 & 34.54 & 26.82 & 41.82 & 23.64 & 48.18 & 22.7  & 16.7  & 15.4  & \textbf{73.4} \\
    Driller & 43.7  & -     & 52.4  & 80.3  & 97.79 & 91.61 & 98.67 & 93.80 & \textbf{99.11} & 44.7  & 65.7  & 62.0  & 92.8 \\
    Duck  & 23.4  & -     & 22.4  & 54.3  & 76.95 & 67.44 & \textbf{87.01} & 72.27 & 85.94 & 15.0  & 25.2  & 37.8  & 62.8 \\
    Eggbox & 22.7  & -     & 21.0  & 34.19 & 44.32 & 41.54 & 43.08 & 39.07 & 48.45 & 25.2  & 50.2  & 70.9  & \textbf{95.3} \\
    Glue  & 43.0  & -     & 50.2  & 67.3  & 80.92 & 66.15 & 75.57 & 66.92 & 74.81 & 32.4  & 49.6  & 65.4  & \textbf{92.5} \\
    Holep. & 12.9  & -     & 15.4  & 45.62 & 74.08 & 60.27 & \textbf{84.51} & 63.30 & 83.50 & 49.5  & 39.7  & 46.9  & 76.7 \\
    \midrule
    Mean  & 27.0  & 27.5  & 28.9  & 48.42 & 65.22 & 60.41 & 74.82 & 61.64 & 75.93 & 32.0  & 40.8  & 47.3  & \textbf{79.2} \\
    \end{tabular}
  }
  \vspace{-2.0em}
\end{table*}%

The proposed refinement procedure has a certain similarity to the point-to-point ICP as well as to PnP. In terms of the ICP, point-to-point correspondences are established by rendering NOCS maps and using the predicted NOCS maps with the same spatial location in 2D, as opposed to the nearest neighbor search in the standard 3D ICP. The backpropagation through the renderer corresponds directly to the distance minimization step of the ICP. However, projective transformation is essential. This is because if only RGB information is used, and therefore predicted NOCS maps in all frames are already in the model's coordinate system and do not impose any additional spatial 3D constraints.  On the other hand, rendering in order to establish correspondences and minimize the discrepancy directly relates to PnP. The proposed refiner can be seen as a multi-view PnP, where direct pixel-wise error, rather than the reprojection error, is minimized.

\subsection{Autolabeling}
We explore the usefulness of the proposed refiner for the self-annotation of weakly labeled real data, similar to the work of \cite{zakharov2020autolabeling}. We assume access to weakly labeled real images with no pose labels. Again, relative transformations between cameras are needed, which is a weaker assumption than having precises per-object pose annotations.
For the sake of simplicity, we use the weak 2D labels to filter out correct detections, although this is not necessary, as the detections can be filtered out with the epipolar constraints.

The autolabeling pipeline operates as follows. First, the YOLO detector and the DPOD network are trained in the usual way on synthetically generated data. They are then applied to a partition of the real data and detections are filtered out.  The poses are computed using PnP+RANSAC and fed into our multi-view refinement pipeline. Finally, we retrieve the newly estimated labels and use them to build a dataset with real images.  The DPOD part is trained as usual, but the predicted NOCS maps are used instead of the ground truth NOCS. The procedure allows us to achieve the performance at the level of DPOD trained on fully annotated real data. No filtering of pose predictions is performed, as ground truth poses are not available in the given scenario. This results in a few images with bad pose annotations being used for training, but they have no significant negative effect on the final performance. The augmented DPOD trained in the standard way for 240 epochs, and the last checkpoint is then used for evaluation.

\section{Experiments}

In this section, we evaluate our multi-view refinement pipeline on LineMOD~\cite{hinterstoisser2012model}, Occlusion~\cite{brachmann2014learning}, Homebrewed~\cite{Kaskman_2019_ICCV_Workshops} and YCB-V~\cite{xiang2018posecnn} datasets to assess its properties. We then evaluate the proposed autolabeling pipeline. Lastly, we test the robustness of the proposed refiner to the imprecision of relative camera transformations and the choice of frames for refinement. We follow the standard evaluation procedure of~\cite{kehl2017ssd,manhardt2018deep,zakharov2019dpod} and report the pose accuracy only for objects correctly detected in 2D on Linemod and Occlusion. Pose quality is computed in accordance with the ADD metric~\cite{hinterstoisser2012model}. On Homebrewed and YCB-V, on the other hand, we submit the predicted poses to the BOP challenge~\cite{bopchallenge} for evaluation and report the Average Recall (AR) metric returned by it.

\textbf{Single object pose estimation.}  Here, we compare the quality of poses predicted with our method to different approaches on the LineMOD~\cite{hinterstoisser2012model} dataset. The results are summarized in Table~\ref{tab:lm}. The table compares our refinement method to various top-performing deep learning methods which reported an ADD score with different pose refinement approaches.  It is not our aim to make direct comparisons with monocular methods, but rather to compare the quality of poses after various refinement methods. 

Our augmented DPOD, labeled as OURS, achieves 53.55\% of the average ADD pose accuracy without any refinement, slightly outperforming the original DPOD, which showed 50\% pose accuracy. Next, we evaluate the performance of our differentiable rendering refiner in the monocular scenario (OURS DR1) and compare it to previous state-of-the-art methods based on deep learning (DL): \cite{rad2017bb8,zakharov2019dpod,li2018deepim,manhardt2018deep}. In this case, the pose accuracy increases from 53.55\% to 67.69\%, even though only one frame is used and no additional multi-view constraints. The refiner outperforms all the competitors in that category apart from DeepIM, even though BB8 and DeepIM have the advantage of using real train data. 

The addition of multiple frames introduces spatial geometric constraints that result in a significant performance boost w.r.t. the ADD score when two (OURS DR2) or four (OURS DR4) frames are used. Even though, the ADD score depends on the choice of the frames, i.e. the closest frames corresponding to weaker constraints, even the two-view refiner can outperform or perform similarly to other approaches using depth-based ICP (\cite{sundermeyer2018implicit} and \cite{kehl2017ssd}) and DenseFusion, which uses real train data, RGBD inference and iterative DL-based refinement. The table clearly shows that even a minimum multi-view setup can bring a significant performance boost without any need for precise calibrated depth information. 

\textbf{Robustness to occlusions.} The robustness of our method to occlusions was evaluated on Occlusion~\cite{brachmann2014learning}, Homebrewed~\cite{Kaskman_2019_ICCV_Workshops} and YCB-V~\cite{xiang2018posecnn} datasets. The results are presented in Table~\ref{tab:occlusion}, Table~\ref{tab:result_hbd} and Table~\ref{tab:result_ycb} respectively. 

Unfortunately, detectors trained on synthetic data are seldom evaluated on Occlusion (Table~\ref{tab:occlusion}). To the best of our knowledge, there is only an ADD score from SSD6D with the DL refinement~\cite{manhardt2018deep}. Therefore, we instead compare to the approaches trained with real data.  The advantage of these approaches is that they use real data from the target domain and also overfit to the particular occlusions present in the test images. This enables comparison with the best detectors and assessment of how closely the refiner is able to approach the performance of the algorithms trained on real data. As can be seen in Table~\ref{tab:occlusion}, our refiner is able to significantly improve the performance of the non-refined baseline. If 4 views are used for refinement, the performance of our approach is on-par with the state-of-the-art results, even though we do not use any real data annotations.

With the Homebrewed dataset~\cite{Kaskman_2019_ICCV_Workshops} (Table~\ref{tab:result_hbd}), all top-performing methods were trained on the synthetic PBR images.Our main aim here is to compare our refiner to CosyPose~\cite{labbe2020}, which also utilizes multi-view refinement. It is clear from the table that correspondence-based pose estimation methods (ours and CDPN) confidently outperform the direct pose prediction of CosyPose if no refinement is used. With the multi-view refinement, the proposed methods achieves top results even if only 2 closest views are used for refinement. Additionally, the proposed approach outperforms CDPN, CosyPose and Pix2Pose even if their predictions are refined with ICP.

Table~\ref{tab:result_ycb} shows a comparison of various top-performing methods on the YCB-V~\cite{xiang2018posecnn} dataset. The dataset comes with a real train set and a set of pre-rendered synthetic images (marked as 'real' and 'synt' in the table). On the other hand, synthetic PBR images are also available. When trained on PBR images, raw non-refined CosyPose poses outperform ours. With refinement, our methods outperforms CosyPose, CDPN and EPOS trained on the same PBR data. However, training on real data still has a huge advantage on this dataset.

\setlength{\tabcolsep}{5pt}
\begin{table}[t]
  \centering
  \caption{Results on the Homebrewed dataset~\cite{Kaskman_2019_ICCV_Workshops} reported according to the Average Recall (AR) metric of the BOP challenge~\cite{bopchallenge} on the BOP challenge subset of test images. \label{tab:result_hbd}}
    
    \begin{tabular}{c|cccc}
    \textbf{Method} & \textbf{Train data} & \textbf{Refinement} & \textbf{AR} & \textbf{Time (s)} \\
    \midrule
    \textbf{OURS} & PBR & 4 furthest views & 0.841 & 0.578 \\
    \textbf{OURS} & PBR & 4 random  views & 0.835 & 0.584 \\
    \textbf{OURS} & PBR & 4 closest views & 0.83  & 0.600 \\
    \textbf{OURS} & PBR & 2 random views & 0.818 & 0.926 \\
    \textbf{OURS} & PBR & 2 furthest views & 0.817 & 0.926 \\
    \textbf{OURS} &  PBR & 2 closest views & 0.787 & 0.91 \\
    CosyPose\cite{labbe2020} & PBR &  8 views & 0.746 & 0.427 \\
    \textbf{OURS} & PBR & no    & 0.725 & 0.163 \\
    CDPNv2\cite{li2019cdpn} & PBR & no    & 0.722 & 0.273 \\
    CosyPose\cite{labbe2020} & PBR & ICP   & 0.712 & 5.326 \\
    CDPNv2 & synt, PBR & ICP   & 0.712 & 0.713 \\
    CosyPose\cite{labbe2020} & PBR & 4 views & 0.696 & 0.445 \\
    Pix2Pose\cite{park2019pix2pose} & PBR & ICP   & 0.695 & 3.248 \\
    CosyPose\cite{labbe2020} & PBR & no    & 0.656 & 0.417 \\
    \end{tabular}%
    \vspace{-2em}
\end{table}%

\textbf{Runtime.} To allow faster refinement time, the camera intrinsics were re-computed such that the rendered image always has the dimensions 128$\times$128.  Moreover, each model was sub-sampled to  contain only 1000 faces. After performing an ablation study on the Linemod dataset, we set the number of refinement iterations to 50. As a result, if 2 views are used for refinement, the average per-image refinement time for one object is 170 milliseconds; with 4 views - 100 milliseconds. Runtime scales linearly with the number of objects in the scene. Tables~\ref{tab:lm},\ref{tab:result_hbd}, \ref{tab:result_ycb} indicate that the proposed refinement, while not being real-time-capable, displays a  performance that is similar in terms of time to the multi-view matching and refinement of CosyPose and similar to or faster than ICP-based methods, depending on the dataset.  All experiments were conducted on an Intel Core i9-9900K CPU 3.60GHz with NVIDIA Geforce RTX 2080 TI GPU.

\textbf{Robustness to camera choices.} For all datasets used for evaluation, we split each sequence into sets of 2 or 4 images. Images in the different sets do not overlap. For each set, we compute relative camera poses from individual ground truth camera poses. The pose is optimized jointly for images in the set. We experiment with three different view sampling strategies: closest views, random views and furthest views.  Tables ~\ref{tab:lm},~\ref{tab:occlusion},~\ref{tab:result_hbd},~\ref{tab:result_ycb}show a quantitative comparison of these strategies. It is clear from the tables that even though close camera locations essentially constitute a bad setup for multi-view pose refinement, as the pose error in one frame is not necessarily visible in other frames, the refiner still improves the poses.  Random and furthest view selections tend to perform similarly on all the datasets.

\textbf{Relative Pose Noise.} The aim of this experiment is to demonstrate how noise in relative poses affects the overall performance of the multi-view optimization pipeline. To do this, we add noise separately to translation and rotation. For rotation transformation, we sample the perturbation angle for each axis from a normal distribution with zero mean and a standard deviation of 5, 7.5 or 10 degrees. For translation transformation, perturbations are sampled from the normal distribution with zero mean and a standard deviation computed on the basis of an object diameter: $\sigma = \frac{1}{3} \cdot (0.1 \cdot diam_{obj})$. Larger deviations render the problem innately ill-posed due to the definition of the ADD measure. If the camera is at a distance of more than 10\% of the model's diameter, the pose will always be classified as incorrect according to the ADD metric. The results can be seen in Table~\ref{tab:noise}. Even with the noisy poses, the refiner still ensures a reasonable pose quality. This table also shows that having more views is beneficial in the noisy scenario.

\setlength{\tabcolsep}{5pt}
\begin{table}[t]
  \centering
  \caption{Results on the YCB-V dataset reported according to the Average Recall (AR) metric of the BOP challenge~\cite{bopchallenge} on the BOP challenge subset of test images. CosyPose~\cite{labbe2020} results labeled with * were obtained by re-running the official implementation of the paper. \label{tab:result_ycb}}
    
    \begin{tabular}{c|cccc}
    \textbf{Method} & \textbf{Train data} & \textbf{Refinement} & \textbf{AR} & \textbf{Time (s)} \\
    \midrule
    CosyPose\cite{labbe2020} & synt + real & ICP   & 0.861 & 2.736 \\
    CosyPose\cite{labbe2020} & synt + real &  8 views & 0.853 & 0.285 \\
    CosyPose\cite{labbe2020} & synt + real & 4 views & 0.84  & 0.318 \\
    CosyPose\cite{labbe2020} & synt + real & no    & 0.821 & 0.241 \\
    Pix2Pose\cite{park2019pix2pose} & synt + real & ICP   & 0.78  & 2.59 \\
    EPOS\cite{Hodan_2020_CVPR}  & synt  & no    & 0.696 & 0.572 \\
    \textbf{OURS} & PBR & 4 furthest  views & 0.674 & 0.555 \\
    \textbf{OURS} & PBR & 4 random  views & 0.645 & 0.557 \\
    \textbf{OURS} & PBR & 2 furthest  views & 0.621 & 848 \\
    \textbf{OURS} & PBR & 2 random views & 0.61  & 0.855 \\
    CosyPose*\cite{labbe2020} & PBR & 8 views & 0.61  & 0.412 \\
    CosyPose*\cite{labbe2020} & PBR & 4 views & 0.592 & 0.415 \\
    CosyPose\cite{labbe2020} &  PBR & no    & 0.574 & 0.342 \\
    \textbf{OURS} & PBR & 4 closest  views & 0.564 & 0.564 \\
    \textbf{OURS} & PBR & 2 closest  views & 0.541 & 0.853 \\
    CDPNv2\cite{li2019cdpn} & synt + real & no    & 0.532 & 0.143 \\
    CDPNv2\cite{li2019cdpn}  & PBR & ICP   & 0.532 & 1.034 \\
    \textbf{OURS} & PBR & no    & 0.525 & 0.187 \\
    EPOS\cite{Hodan_2020_CVPR}  & PBR & no    & 0.499 & 0.764 \\
    CDPNv2\cite{li2019cdpn} & PBR & no    & 0.39  & 0.448 \\
    \end{tabular}%
     \vspace{-2em}
\end{table}%

\textbf{Autolabeling.} In this experiment, we retrieve the estimated OURS+DR2 and OURS+DR4 labels and use them to replace the synthetic training set by the automatically generated real labels as discussed above. The extended training set is then used to fine-tune the synthetically trained baseline. The results can be seen in Table~\ref{tab:autolabel}. It can be clearly seen that the fine-tuned network trained on autolabels significantly outperforms the synthetic DPOD and is very competitive when compared to the state-of-the-art methods trained on full ground truth labels.

\setlength{\tabcolsep}{4pt}
\begin{table}[h]
  \centering
  \caption{Percentages of correctly estimated poses w.r.t. the ADD score on the Linemod~\cite{hinterstoisser2012model} dataset with noisy relative camera transformations.
  \label{tab:noise}}
\resizebox{1\linewidth}{!}{%
    \begin{tabular}{c|ccc|ccc|ccc|ccc|}
          & \multicolumn{6}{c|}{2 views}                  & \multicolumn{6}{c|}{4 views} \\
\cmidrule{2-13}          & \multicolumn{3}{c|}{Trans. (\% diam.)} & \multicolumn{3}{c|}{Rot. (deg.)} & \multicolumn{3}{c|}{Trans. (\% diam.)} & \multicolumn{3}{c|}{Rot. (deg.)} \\
\cmidrule{2-13}          & 5     & 10    & 15    & 5     & 7.5   & 10    & 5     & 10    & 15    & 5     & 7.5   & 10 \\
    \midrule
    Ape   & 95  & 79 & 58 & 93 & 82 & 68 & 99 & 89 & 66 & 98 & 89 & 73 \\
    Bvise & 97 & 85 & 63 & 97 & 91 & 78 & 100   & 91 & 70 & 99 & 93 & 80 \\
    Cam   & 97 & 86 & 65 & 95 & 78 & 65  & 100   & 93 & 56    & 99 & 86 & 66 \\
    Can   & 87 & 75 & 53 & 86 & 73 & 63 & 89 & 81  & 56    & 90 & 80 & 62 \\
    Cat   & 95 & 81 & 62 & 95 & 85 & 72 & 99  & 89    & 63  & 99 & 91  & 76 \\
    Driller & 95 & 83 & 61 & 94 & 85 & 67 & 99 & 91 & 68 & 99    & 91 & 75 \\
    Duck  & 89 & 74 & 57 & 91 & 78 & 60 & 97 & 86 & 62 & 96 & 81 & 63 \\
    Eggbox & 97 & 96  & 94 & 97 & 98 & 96 & 99 & 98 & 98 & 98 & 98 & 98 \\
    Glue  & 94 & 93 & 86 & 94 & 94  & 91 & 98 & 96  & 96 & 97 & 99 & 97 \\
    Holep. & 79 & 65 & 44 & 73 & 63 & 51 & 89 & 76 & 52  & 85 & 73 & 62 \\
    Iron  & 96 & 79 & 60 & 97 & 87 & 75 & 99 & 87 & 62 & 99 & 93 & 78 \\
    Lamp  & 92 & 80 & 61 & 91 & 79 & 68 & 96 & 88 & 65 & 96  & 91 & 77 \\
    Phone & 95 & 79 & 60 & 95 & 86 & 71 & 100   & 87  & 60 & 99 & 87  & 72 \\
    \midrule
    Mean  & 93 & 81 & 63 & 92 & 83 & 71 & 97 & 89 & 67 & 96 & 89 & 75 \\
    \end{tabular}%
    }
\end{table}%

\setlength{\tabcolsep}{3pt}
\begin{table}[h]
  \centering
  \caption{Percentages of correctly estimated poses w.r.t. the ADD score when automatically annotated data is used to train the monocular detector.
  \label{tab:autolabel}}
  \resizebox{1\linewidth}{!}{%
    \begin{tabular}{c|c|cc|ccccc}
    \multirow{2}[3]{*}{\textbf{Train data}} & \multirow{2}[3]{*}{\textbf{Synt}} & \multicolumn{2}{c|}{\textbf{Real - Autolabels}} & \multicolumn{5}{c}{\multirow{2}[3]{*}{\textbf{Real - GT labels}}} \\
\cmidrule{3-4}          &       & \textbf{DR2} & \textbf{DR4} & \multicolumn{5}{c}{} \\
    \midrule
    \textbf{Method} & \textbf{OURS} & \textbf{OURS} & \textbf{OURS} & \textbf{DPOD\cite{zakharov2019dpod}} & \textbf{PVNet\cite{peng2019pvnet}} & \textbf{OURS} & \textbf{CDPN\cite{li2019cdpn}} & \textbf{HybP.\cite{song2020hybridpose}} \\ \midrule
    Ape   & 28.04 & 65.29 & 68.84 & 53.28 & 43.62 & 74.78 & 64.38 & 77.6 \\
    Bvise & 71.65 & 98.84 & 99.81 & 95.34 & 99.90 & 99.71 & 97.77 & 99.6 \\
    Cam   & 58.62 & 93.19 & 94.25 & 90.36 & 86.86 & 96.52 & 91.67 & 95.9 \\
    Can   & 62.80 & 97.44 & 98.62 & 94.1  & 95.47 & 99.02 & 95.87 & 93.6 \\
    Cat   & 40.12 & 78.54 & 82.04 & 60.38 & 79.34 & 90.52 & 83.83 & 93.5 \\
    Driller & 61.46 & 94.81 & 96.47 & 97.72 & 96.43 & 97.9  & 96.23 & 97.2 \\
    Duck  & 26.17 & 31.05 & 32.46 & 66.01 & 52.58 & 48.41 & 66.76 & 87 \\
    Eggbox & 87.65 & 93.48 & 95.24 & 99.72 & 99.15 & 99.9  & 99.72 & 99.6 \\
    Glue  & 67.48 & 95.26 & 96.91 & 93.83 & 95.66 & 97.68 & 99.61 & 98.7 \\
    Holep. & 19.09 & 25.43 & 43.47 & 65.83 & 81.92 & 49.62 & 85.82 & 92.5 \\
    Iron  & 83.85 & 99.26 & 99.59 & 99.8  & 98.88 & 99.39 & 97.85 & 98.1 \\
    Lamp  & 43.52 & 90.79 & 93.31 & 88.11 & 99.33 & 92.92 & 97.89 & 96.9 \\
    Phone & 45.77 & 88.14 & 90.65 & 74.24 & 92.41 & 93.07 & 90.75 & 98.3 \\
    \midrule
    Mean  & 53.55 & 80.89 & 83.97 & 82.98 & 86.27 & 87.65 & 89.86 & 94.5 \\
    \end{tabular}%
  }
  
\end{table}%

\section{Conclusions}

In this paper, we propose a novel object pose refinement pipeline. We adopt the idea of using multiple frames to produce a single joint  object pose estimate. The use of multiple frames enables effective use of geometric constraints.   The proposed refiner is based on the external object detector, which outputs deep 2D-3D correspondences in the form of Normalized Object Coordinate space. Predicted dense correspondences from calibrated multiple frames can be brought together to obtain a better pose estimate using a differentiable renderer. The proposed approach imposes no constraints on the number of frames used and the exact positions of the cameras in the 3D world, and it remains robust even when the cameras are in  close proximity to one another. 
We experimentally show that the refiner works excellently on Linemod, Occlusion, Homebrewed and YCB-V datasets.
As an alternative use case, we demonstrate how the proposed approach can be applied to automatically annotate real train data, which has no object pose annotations.  
The approach even remains effective if the relative transformations in-between frames are imprecise. 
Moreover, even a multi-view setup with only two frames already produces excellent results. This demonstrates that the approach can be used successfully in practical applications.



%
\bibliographystyle{IEEEtran}
\bibliography{egbib}




\end{document}